\documentclass[conference]{IEEEtran}
\IEEEoverridecommandlockouts

\usepackage{cite}
\usepackage{amsmath,amssymb,amsfonts}
\usepackage{algorithmic}
\usepackage{graphicx}
\usepackage{textcomp}
\usepackage{subfig}
\usepackage{subcaption}
\usepackage{comment}

\usepackage{xcolor}
\def\BibTeX{{\rm B\kern-.05em{\sc i\kern-.025em b}\kern-.08em
    T\kern-.1667em\lower.7ex\hbox{E}\kern-.125emX}}

\makeatletter

\def\ps@IEEEtitlepagestyle{
  \def\@oddfoot{\mycopyrightnotice}
  \def\@evenfoot{}
}
\def\mycopyrightnotice{
  {\footnotesize 979-8-3503-9431-3/23/\$31.00 ~\copyright~2023 IEEE\hfill} 
  \gdef\mycopyrightnotice{}
}

\@ifundefined{showcaptionsetup}{}{
 \PassOptionsToPackage{caption=false}{subfig}}
\usepackage{subfig}
\makeatother

\usepackage{eso-pic}
\newcommand\AtPageUpperMyright[1]{\AtPageUpperLeft{
 \put(\LenToUnit{0.5\paperwidth},\LenToUnit{-1cm}){
     \parbox{0.5\textwidth}{\raggedleft\fontsize{9}{11}\selectfont #1}}
 }}
\newcommand{\conf}[1]{
\AddToShipoutPictureBG*{
\AtPageUpperMyright{#1}
}
}

\begin{document}

\title{Investigation of Polycystic Ovary Syndrome (PCOS)
Diagnosis Using Machine Learning Approaches}
\conf{2023 5\textsuperscript{th} International Conference on Sustainable Technologies for Industry 5.0 (STI), 09-10 December, Dhaka}

\title{Investigation of Polycystic Ovary Syndrome (PCOS)
Diagnosis Using Machine Learning Approaches}

\author{\IEEEauthorblockN{Al Zadid Sultan Bin Habib\textsuperscript{1}, Md Asif Bin Syed\textsuperscript{2}, Md. Ekramul Islam\textsuperscript{3}, Tanpia Tasnim\textsuperscript{4}}
  \IEEEauthorblockA{
    \textsuperscript{1}Lane Department of Computer Science and Electrical Engineering, West Virginia University, Morgantown, WV 26506, USA\\
    \textsuperscript{2}Department of Industrial and Management Systems Engineering, West Virginia University, Morgantown, WV 26506, USA\\
    \textsuperscript{3}Department of Computer Science \& Engineering, Stamford University Bangladesh, Dhaka-1217, Bangladesh\\
    \textsuperscript{4}Department of Computer Science and Engineering, Green University of Bangladesh, Narayanganj-1461, Dhaka, Bangladesh\\
    Email: ah00069@mix.wvu.edu\textsuperscript{1}, ms00110@mix.wvu.edu\textsuperscript{2}, eislam706@gmail.com\textsuperscript{3}, tanpia@cse.green.edu.bd\textsuperscript{4}
  }
}

\maketitle

\begin{abstract}
Polycystic Ovarian Syndrome (PCOS) is a widespread hormone problem for women of childbearing age. Women with PCOS may not ovulate; they might have high levels of androgens and have many small cysts on the ovaries. It can cause missed or irregular menstrual periods, excess hair growth, acne, infertility, and weight gain. Machine Learning (ML) can effectively diagnose this disease at an earlier stage as tons of medical data are available now. Traditional approaches to detect PCOS encompass a combination of clinical evaluation, medical history assessment, physical examination, and laboratory tests. These approaches aim to identify the characteristic symptoms and hormonal imbalances associated with PCOS. Physical examination requires good resources and costs time and money. In recent times, data-driven techniques have substantially advanced disease prediction within the medical field. We aim to utilize ML approaches, incorporating unique feature selection algorithms, to predict PCOS. This paper introduces a data-driven approach to PCOS diagnosis, combining Feature Engineering and ML. Several feature selection approaches have been considered to select sets of features for training the ML model, including CatBoost, Extreme Gradient Boosting (XGBoost), Light Gradient Boosting Machine (LGBM), AdaBoost, Random Forest (RF). Results demonstrate that AdaBoost, with ten features selected by RF Feature Importance and Highest Correlation (HC), provides the highest test accuracy.
\end{abstract}

\begin{IEEEkeywords}
PCOS, CatBoost, Machine Learning, Random
Forest Feature Importance, Feature Selection.
\end{IEEEkeywords}

\section{Introduction and Related Works}
PCOS is a chronic endocrine disorder in females with increased androgen levels. Treatment for PCOS is focused on a few key characteristics, such as menstruation disruption, anovulation, and symptoms of hyperandrogenism \cite{b1}. In women of reproductive age (15-49 years), this issue affects 5–10\% of them, with a reported prevalence of 8\% in African American women and 4.8\% in white women \cite{b2}, \cite{b3}. Women with this ovarian malfunction are more likely to have high blood pressure, obesity, gynecological cancer, type 2 diabetes, and other conditions. Additionally, new studies have demonstrated that PCOS increases the chance of miscarriage in the first trimester, as it causes follicles to grow improperly in the ovaries, where they are stopped at an early stage and miscarry before maturing, contributing to infertility. To avoid any uncomfortable symptoms of PCOS condition, it is essential to test patients as early as possible.

ML techniques have shown promise in predicting and detecting PCOS. Researchers have utilized various ML algorithms and approaches to develop prediction models for PCOS. One study proposed an extended ML classification technique for PCOS prediction using ovary ultrasound images. The study employed a Convolutional Neural Network (CNN) for feature extraction from the images and used stacking ensemble ML techniques for classification. The results demonstrated the potential of ML in accurately classifying between PCOS and non-PCOS ovaries \cite{b4}. Particle Swarm Optimization (PSO) is used in the work of \cite{b5} to segment the follicles while modifying the fitness function. Three classifiers: the Support Vector Machine (SVM) with RBF kernel, the k-nearest Neighbors (kNN) with Euclidean distance, and the Neural Network (NN) with learning vector quantization are used in the research by the authors in \cite{b6}.

In addition to utilizing stereology, researchers have pinpointed the presence of the polycystic ovary. This method involves assessing the number and dimensions of individual follicles. Additionally, the follicle diameter is computed using the Euclidean distance approach \cite{b7}. In contrast, the Gabor Wavelet is utilized for feature extraction in \cite{b8}, \cite{b9}. In addition, backpropagation is modified by Conjugate Gradient Fletcher Reeves and Lavenberg-Marquardt Optimization to categorize PCOS. Follicles are divided into sections in \cite{b10} using a region-growing system. This method determines if the initial seeds’ neighbor should be included in the segmentation region. According to the authors of \cite{b11}, a median filter could reduce noise in PCOS pictures. This filter’s primary goal is to locate a median in a particular picture element window. A window’s median will be updated in the main window. The Otsu global threshold \cite{b12} of the image can be used to determine how similar the two pixels are. The threshold values are iterated through Otsu’s thresholding approach. This threshold also calculates the spread for the pixel levels on either side of the threshold. A computational approach to Canny edge detection was proposed in \cite{b13} for the Canny edge detection to diagnose PCOS, follicles can be seen on the ultrasound scan.

However, there needs to be more research on using ML techniques in clinical aspects of PCOS patient screening. So, the main contribution of this paper is to select significant features resulting in PCOS using the art of state feature selection algorithms. Furthermore, it introduces a unique combination of feature selection algorithms along with ML techniques that require minimum preprocessing for finding the best prediction results. This predictive capability allows for early interventions and lifestyle modifications, aligning with Industry 5.0's goal of preventive healthcare. Also, Industry 5.0 places a strong emphasis on data-driven decision-making for Healthcare 5.0. The data-driven approach has significantly improved predictions in medical science for disease detection. In traditional PCOS detection, healthcare providers assess the patient's medical history, including menstrual irregularities, signs of hyperandrogenism (such as hirsutism and acne), and family history. They also conduct physical exams, which involve checking BMI, blood pressure, and signs like excess hair growth or acne to identify ovarian issues. Contributions include selecting significant PCOS-related features using advanced feature selection algorithms and introducing a unique combination of feature selection and ML techniques with minimal preprocessing for improved prediction results. In section II, the methodological frameworks are discussed in detail, and section III introduces the used feature selection approaches. Results and discussions are analyzed in section IV, and the paper is concluded in section V.

\section{Methodological Framework}
The following subsections can be denoted as the methodological framework for this work, starting from Input Dataset to Performance Evaluation.

\subsection{Input Dataset}
The dataset referenced as \cite{b14} is a comprehensive collection of physical and clinical parameters crucial for identifying PCOS and infertility-related issues. It was gathered from ten medical facilities in Kerala, India, and is now publicly available. This dataset includes data from 541 patients and encompasses 43 distinct features, along with a single target attribute categorized into binary classes: `1' indicating the presence of PCOS and `0' denoting the absence of PCOS. To enable analysis, the dataset was imported into a Pandas data frame.

\subsection{Missing Values Imputation}
\begin{figure}
\centerline{\includegraphics[scale=0.20]{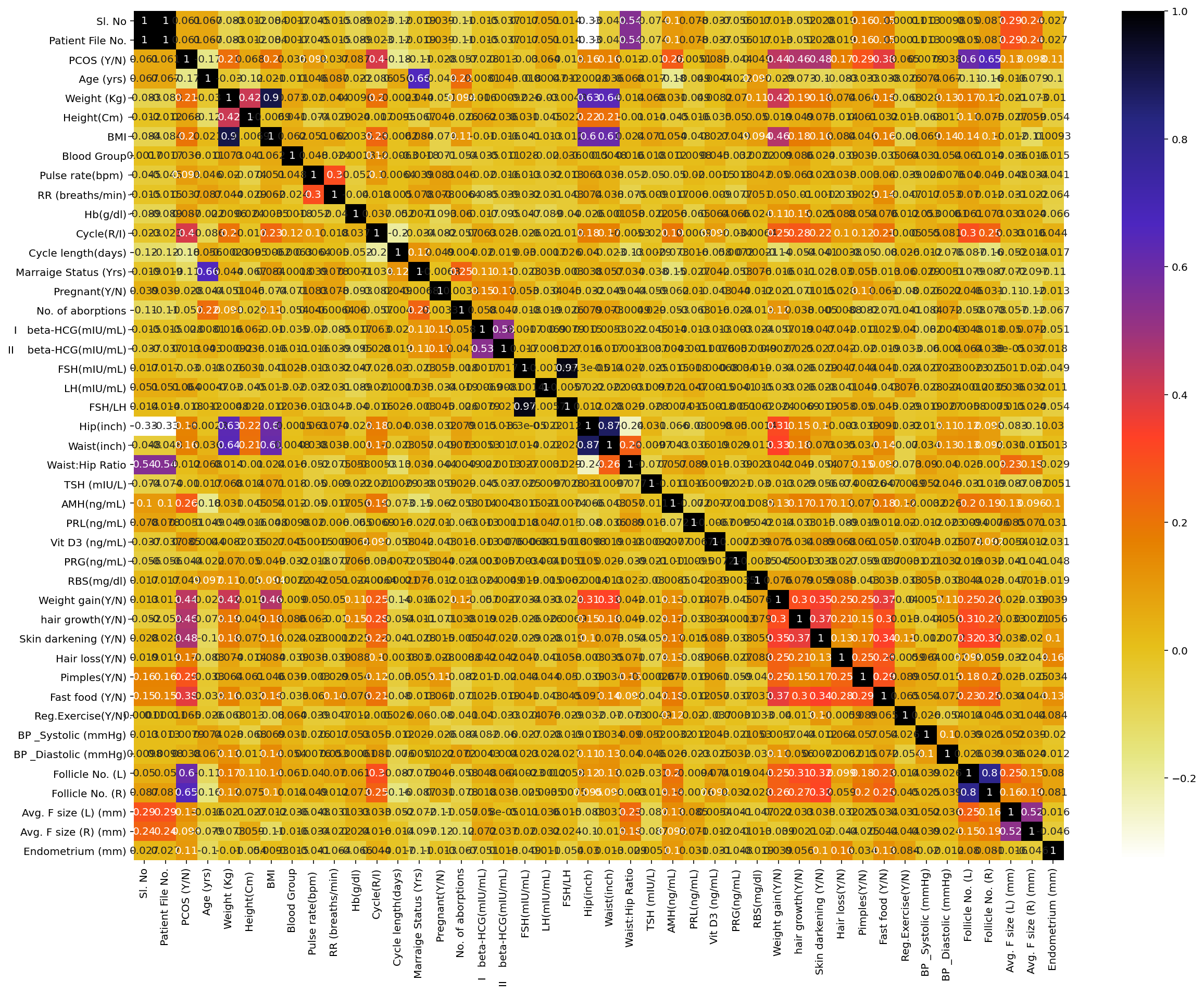}}
\caption{Heatmap correlation matrix for the dataset.}
\label{fig1}
\end{figure}

Linear imputation, a commonly used method for addressing missing values, involves replacing them with the mean values of the respective columns \cite{b15}. In this approach, null values were substituted with the column means. Fig.~\ref{fig1} illustrates the heatmap correlation matrix.

\subsection{Normalization}
Min-max normalization is a commonly used data normalization method, where the minimum and maximum values of each feature are transformed to 0 and 1, respectively. Simultaneously, all other values within each feature are proportionally scaled to decimal values within the range of 0 to 1 \cite{b16}.

\subsection{Sampling}
\begin{figure}
\centerline{\includegraphics[scale=0.60]{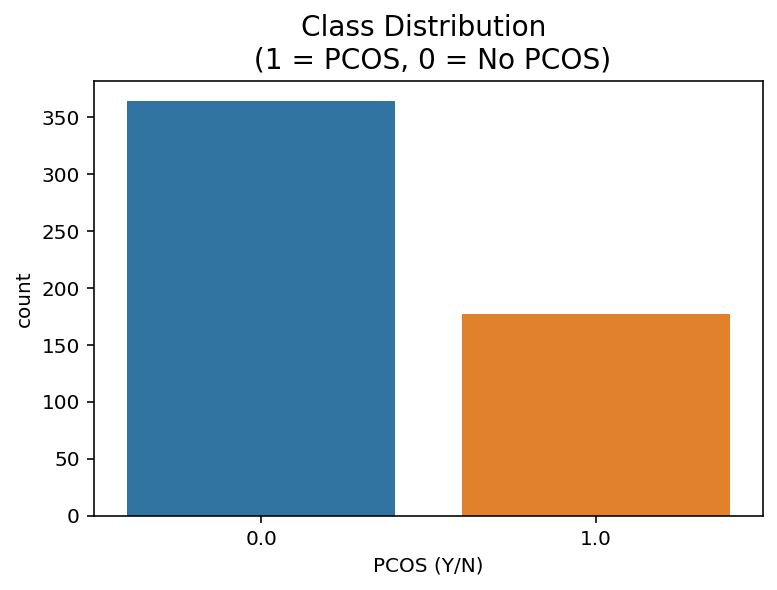}}
\caption{Original class distribution of the dataset.}
\label{fig2}
\end{figure}

To address the data imbalance issue in our classification problem, where the data classes are not evenly distributed, we employed undersampling and oversampling techniques. Specifically, we utilized random undersampling by randomly selecting examples from the majority class and removing them from the training set until achieving a more balanced distribution \cite{b17}. Additionally, we applied the Synthetic Minority Over-sampling Technique (SMOTE), a widely used method in handling class imbalance. SMOTE generates synthetic examples for the minority class by interpolating between existing samples, effectively increasing the representation of the minority class in the training data. This approach helps prevent bias in our classification model \cite{b18}. Fig.~\ref{fig2} illustrates the class distribution before these sampling techniques were applied.

\subsection{Dataset Splitting}
The subsampled dataset underwent a series of splits to create distinct sets for training, validation, and testing. Initially, the dataset was divided into a 70\%-30\% ratio, yielding a train and subtest split. Subsequently, the subtest dataset was divided into a 50\%-50\% ratio to create separate validation and test sets. Finally, the equally balanced subsampled dataset was categorized into three portions: 70\% allocated for training, 15\% for validation, and 15\% for testing.

\subsection{ML Models}
Our PCOS diagnostic approach relies on ML models, carefully chosen to address the dataset's characteristics and PCOS diagnostic requirements. The selected models include CatBoost, XGBoost, LGBM, AdaBoost, and RF, each chosen for its suitability in tackling the unique challenges of PCOS diagnosis.
\subsubsection{CatBoost}
CatBoost, developed by Yandex, represents a cutting-edge ML algorithm capable of handling diverse data types and collaborating with other algorithms to tackle various everyday challenges. It distinguishes itself by delivering top-tier precision without the extensive data training typically required by other ML models. CatBoost excels in creating effective data designs and exhibits compatibility with various data types, such as audio, text, images, and historical data. This algorithm particularly shines when categorical features play a vital role. While it offers faster prediction times than some algorithms, it demands more training than Gradient Boosting \cite{b19}.

\subsubsection{XGBoost}
XGBoost is a widely used ML algorithm known for its versatility in various supervised learning tasks, including regression, classification, and ranking. Derived from the Gradient Boosting Machine (GBM) framework, XGBoost has been carefully crafted to enhance algorithm performance, pushing the computational boundaries to provide flexible and accurate results \cite{b20}.

\subsubsection{LGBM}
LGBM, an advanced variation of GBM, employs a gradient-boosting framework with a unique approach to learning based on trees. In LGBM, trees grow vertically and horizontally, with one branch expanding in terms of tree leaves and the other in terms of levels. Notably, the leaf-wise algorithm allows for more significant loss reduction when creating similar leaves than the level-wise algorithm. The "Light" prefix reflects its computational efficiency, enabling faster computations, lower memory requirements, and efficient handling of large datasets. Its commitment to accuracy and support for GPU learning make LGBM increasingly popular among data scientists for developing data science applications. However, caution is necessary to avoid overfitting when squeezing extensive information into limited space \cite{b21}.

\subsubsection{AdaBoost}
AdaBoost, one of the pioneering boosting methods, excels at converting multiple weak classifiers into robust ones, making it applicable to various learning algorithms. It combines the output of multiple algorithms through weighted voting, representing the collective output of the boosted algorithm. However, AdaBoost is sensitive to noisy input and outliers. Nonetheless, it exhibits relative resilience to overfitting compared to some other learning algorithms \cite{b22}.

\subsubsection{RF}
RF is another effective and faster-to-optimize ensemble learning method compared to LGBM. RF employs multiple decision trees to form an ensemble. Each tree within the forest contributes to class predictions, and the final prediction is determined by majority voting among the trees once the votes are collected \cite{b23}.

The choice of gradient boosting algorithms (CatBoost, XGBoost, and LGBM) for our study was driven by their capacity to handle intricate, non-linear relationships in the data, which is crucial for PCOS diagnosis. These algorithms are recognized for their robustness against overfitting, a critical consideration when working with medical data prone to noise. Additionally, they excel in feature importance analysis, enabling us to identify the most influential factors in PCOS diagnosis. AdaBoost, another chosen technique, leverages multiple weak learners to enhance diagnostic accuracy, which aligns with the complexity of PCOS diagnosis. Its adaptability to diverse base classifiers makes it suitable for the varied types of features and relationships in our dataset. Finally, RF was included due to its effectiveness in handling high-dimensional datasets with numerous features and its capability to estimate feature importance, aiding in identifying crucial clinical attributes contributing to PCOS diagnosis. Our chosen ML models exhibit limitations, including overfitting, despite mitigation efforts through strategies like cross-validation and hyperparameter tuning. Additionally, specific models, like ensembles, may lack interpretability, posing challenges in medical contexts where explanation is crucial. Furthermore, Deep Learning (DL) models may demand larger datasets than available for PCOS diagnosis, impacting their generalization performance.

\subsection{Train and Test Phase}
The labeled dataset was divided into training and test sets, with the training set initially employed to train all ML models. Subsequently, the models' learning performance was evaluated using the test set.

\subsection{Performance Evaluation}
Various performance evaluation metrics were utilized to assess the models' performance, with test accuracy being a widely recognized and crucial parameter. In addition to test accuracy, the evaluation included metrics such as precision, recall, F1-score, Mathew’s Correlation Coefficient (MCC), and Cohen’s Kappa (CK) scores to provide a comprehensive evaluation of the models' performance.

\section{Feature Selection}
In our study, the careful selection of feature selection methods played a pivotal role in improving the performance of ML models for PCOS diagnosis. We provide a thorough explanation of the rationale behind these chosen methods and elucidate their significant impact on the performance of the ML models. Feature selection plays a crucial role in traditional ML processes, enhancing the predictive capabilities of ML algorithms by selecting essential attributes and removing redundant ones. This practice minimizes the potential for making decisions based on noise and overfitting while improving model accuracy and reducing training time \cite{b24}. Limitations include feature selection assumptions, potential information loss, and dataset issues like imbalance and data quality. Generalizability to diverse populations may be limited, and ethical considerations, such as algorithmic bias, should be noted. Future research should focus on interpretable models and incorporating domain knowledge to address these limitations.

\subsection{Random Forest Feature Importance}
\begin{figure}
\centerline{\includegraphics[scale=0.55]{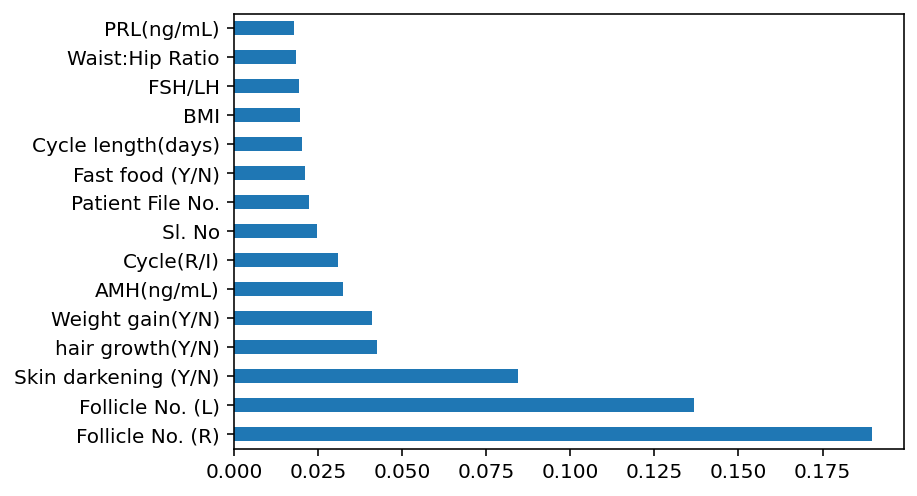}}
\caption{Top 15 features ranked by RF feature importance.}
\label{fig3}
\end{figure}

RF Feature Importance was selected for its robustness in identifying influential features, aiding in feature ranking, and focusing on essential clinical attributes for PCOS diagnosis. This approach improved model accuracy by reducing noise and providing insights into vital physiological factors. RF is known for its ability to provide robust predictive performance while mitigating overfitting and offering ease of interpretability. This is achieved through its straightforward approach to determining the importance of each feature in the decision trees. Typically, RF comprises hundreds of decision trees, each focusing on specific features or combinations. The dataset is divided into two buckets, and the purity of each bucket determines the importance of individual features, often assessed using Gini impurity as the measure of impurity. Features contributing to reducing impurity are considered more influential \cite{b25}. Fig.~\ref{fig3} illustrates the ranking of the top 15 features selected by RF based on their importance.

\subsection{Chi-Squared Test Score}
The Chi-Squared Test Score (Ch2) was chosen to emphasize the importance of categorical features in PCOS diagnosis, ensuring their proper consideration in the ML models. Neglecting categorical data could result in inaccurate predictions, and this method helped maintain the robustness of the diagnostic process by handling categorical variables effectively. In statistics, the Chi-Square test is a tool for assessing potential independence between two events by comparing observed counts (O) and expected counts (E) extracted from a dataset of two variables. The chi-square test quantifies the disparity between the expected and observed counts. When selecting features, the objective is to identify those that have a significant dependence on the outcome. In cases where two features are independent, the observed count closely matches the expected count, resulting in a lower Chi-Square value. Conversely, a high Chi-Square score indicates that the independence hypothesis is likely false. In essence, features exhibiting more substantial reliance on the response variable and yielding larger Chi-Square values are preferred choices for model training \cite{b26}.

\subsection{Highest and Least Correlation Score}
The Correlation Score evaluated feature interdependencies and their impact on the target attribute. It assisted in identifying strong feature correlations, managing multicollinearity, and guiding feature selection, ultimately streamlining the learning process and improving model performance. Statistically, correlation quantifies the linear relationship between two variables, indicating how closely they are related. Therefore, selecting two highly correlated variables provides redundant information to the model. Features with an HC exhibit linear dependence and have nearly identical effects on the dependent variable. By evaluating the relationship between each feature and the target variable, we can identify features with the strongest associations with the target. Additionally, for feature selection, it is crucial to consider features with the Least Correlation (LC), as they may provide unique information and contribute to model diversity and robustness, ultimately enhancing the model's performance. This distinction in correlation values can be computed when selecting features based on the target variable itself \cite{b27}.
\section{Results Analysis}
\subsection{Random Undersampling Minority Class}
\begin{figure}
  \centering
  \subfloat[CatBoost]{\includegraphics[width=0.45\columnwidth]{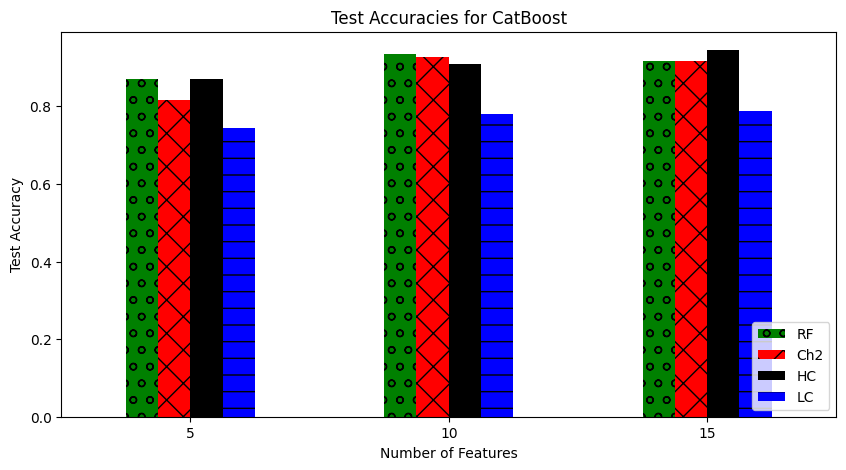}
  \label{fig41}}
  \hfil
  \subfloat[XGBoost]{\includegraphics[width=0.45\columnwidth]{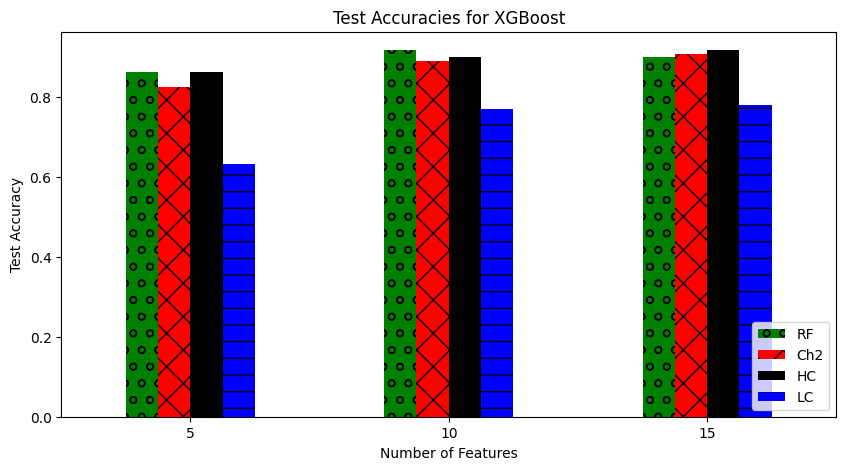}
  \label{fig42}}
  \hfil
  \subfloat[LGBM]{\includegraphics[width=0.45\columnwidth]{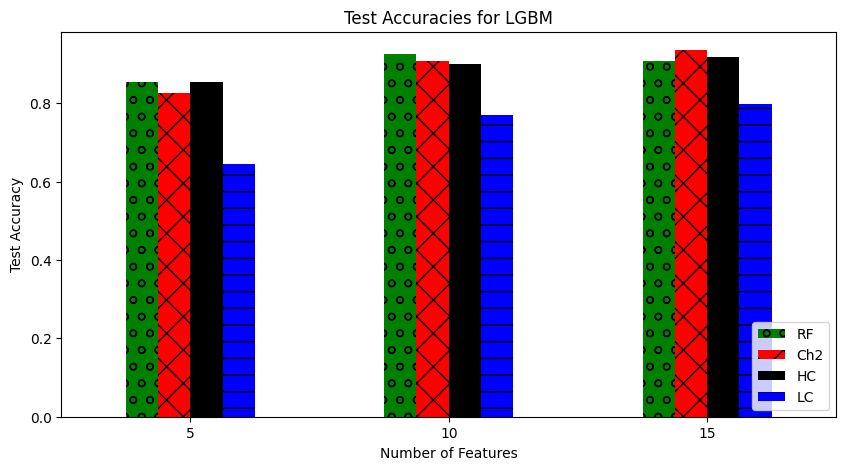}
  \label{fig43}}
  \hfil
  \subfloat[AdaBoost]{\includegraphics[width=0.45\columnwidth]{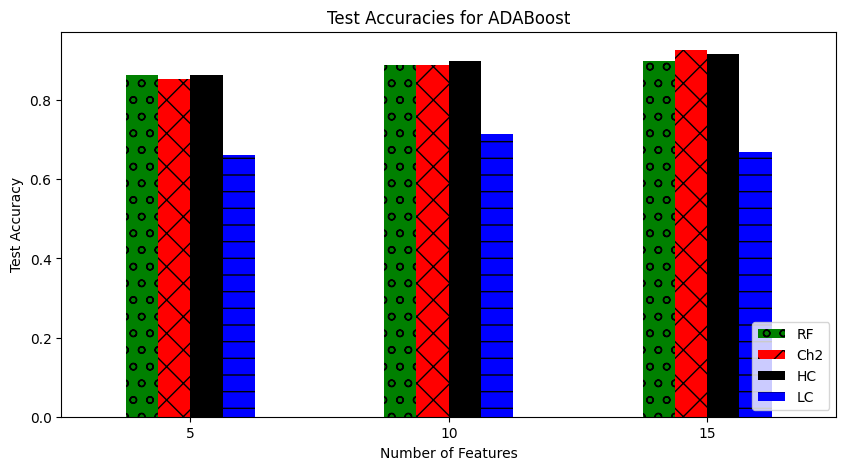}
  \label{fig44}}
  \hfil
  \subfloat[RF]{\includegraphics[width=0.45\columnwidth]{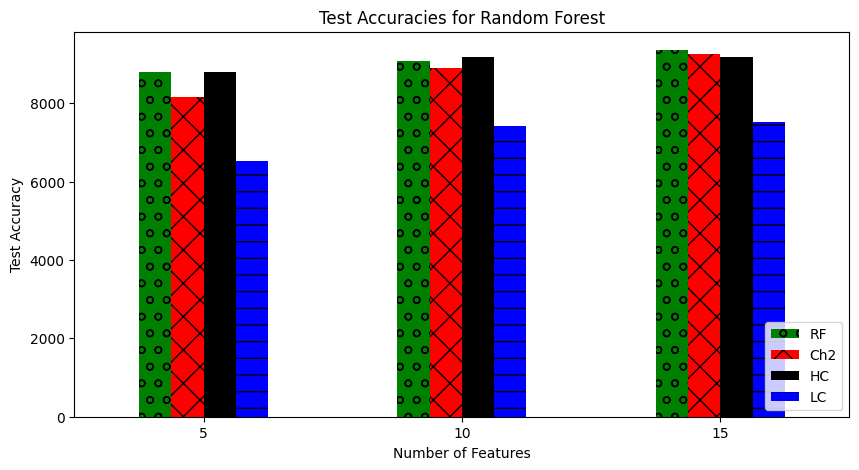}
  \label{fig45}}
  \caption{Test accuracies of undersampling for different ML algorithms.}
  \label{fig4}
\end{figure}

In the evaluation phase, we strategically employed the undersampling technique, a method designed to balance class distributions by randomly eliminating instances from the majority class. This approach's effectiveness is evident in the test accuracy results presented in Table ~\ref{tab1} for CatBoost, where we achieved the highest test accuracy of 90.74\% for both 10 and 15 features selected via RF feature importance. For a visual comparison of these outcomes, we offer a detailed analysis in Fig. ~\ref{fig41}. Furthermore, we present the test accuracies for XGBoost in Table ~\ref{tab1}, accompanied by a visual representation in Fig. ~\ref{fig42}, illustrating the comparison alongside CatBoost.
Of particular note is the consistent achievement of the highest test accuracy at 87.04\% for both 10 and 15 features selected using RF feature importance. This remarkable consistency across models underscores the robustness of the chosen features. In Table ~\ref{tab1}, we provide a comprehensive view of the results for LGBM, which yielded the highest test accuracy of 90.74\% when utilizing 15 features identified by RF feature importance. AdaBoost stands out with the highest test accuracy of 94.44\% for 10 features selected through RF feature importance and HC, as indicated in Table ~\ref{tab1} and Fig. ~\ref{fig44}. Lastly, RF demonstrated superior performance, delivering a test accuracy of 88.89\% for 10 features chosen based on RF feature importance and 15 features selected via HC, as demonstrated in Table ~\ref{tab1} and Fig. ~\ref{fig45}. It is essential to emphasize that these algorithms exhibited exceptional performance, particularly when employing the 10 features selected based on RF feature importance.
\begin{table}
\caption{Test accuracies of different algorithms with different features selection approaches for undersampling}
\begin{center}
\begin{tabular}{|c|c|c|c|c|}
\hline
\textbf{No. of Features} & \textbf{RF} & \textbf{Ch2} & \textbf{HC} & \textbf{LC} \\
\hline
\multicolumn{5}{|c|}{CatBoost}\\
\hline
5 & 83.33\% & 75.93\% & 83.33\% & 53.70\%\\
\hline
10 & 90.74\% & 83.33\% & 87.04\% & 66.67\%\\
\hline
15 & 90.74\% & 88.89\% & 85.19\% & 51.85\%\\
\hline
\multicolumn{5}{|c|}{XGBoost}\\
\hline
5 & 75.93\% & 72.22\% & 75.93\% & 59.26\%\\
\hline
10 & 87.04\% & 81.48\% & 81.48\% & 61.11\%\\
\hline
15 & 87.04\% & 83.33\% & 75.93\% & 59.26\%\\
\hline
\multicolumn{5}{|c|}{LGBM}\\
\hline
5 & 77.78\% & 72.22\% & 77.78\% & 55.56\%\\
\hline
10 & 88.89\% & 83.33\% & 85.19\% & 59.26\%\\
\hline
15 & 90.74\% & 83.33\% & 83.33\% & 57.41\%\\
\hline
\multicolumn{5}{|c|}{AdaBoost}\\
\hline
5 & 90.74\% & 74.07\% & 90.74\% & 53.70\%\\
\hline
10 & 94.44\% & 87.04\% & 94.44\% & 50.00\%\\
\hline
15 & 87.04\% & 87.04\% & 81.48\% & 48.15\%\\
\hline
\multicolumn{5}{|c|}{RF}\\
\hline
5 & 81.48\% & 75.93\% & 81.48\% & 61.11\%\\
\hline
10 & 88.89\% & 85.19\% & 87.04\% & 61.11\%\\
\hline
15 & 85.19\% & 85.19\% & 88.89\% & 62.96\%\\
\hline
\end{tabular}
\end{center}
\label{tab1}
\end{table}
\begin{figure}
  \centering
  \subfloat[CatBoost]{\includegraphics[width=0.45\columnwidth]{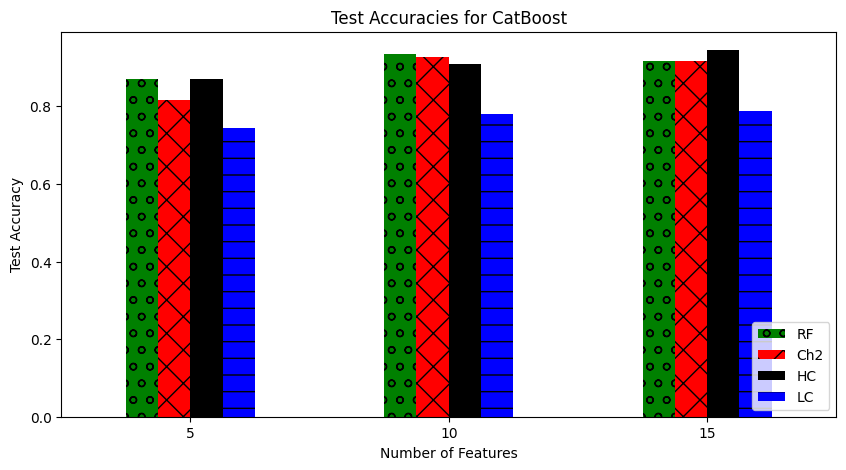}
  \label{fig51}}
  \hfil
  \subfloat[XGBoost]{\includegraphics[width=0.45\columnwidth]{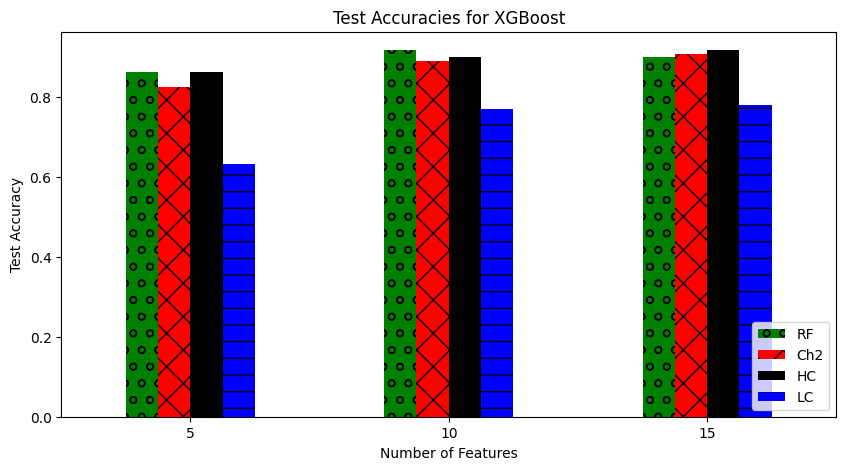}
  \label{fig52}}
  \hfil
  \subfloat[LGBM]{\includegraphics[width=0.45\columnwidth]{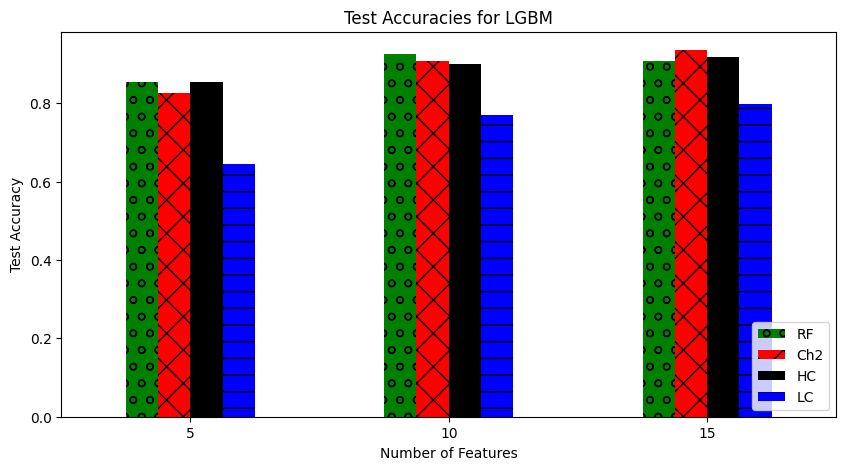}
  \label{fig53}}
  \hfil
  \subfloat[AdaBoost]{\includegraphics[width=0.45\columnwidth]{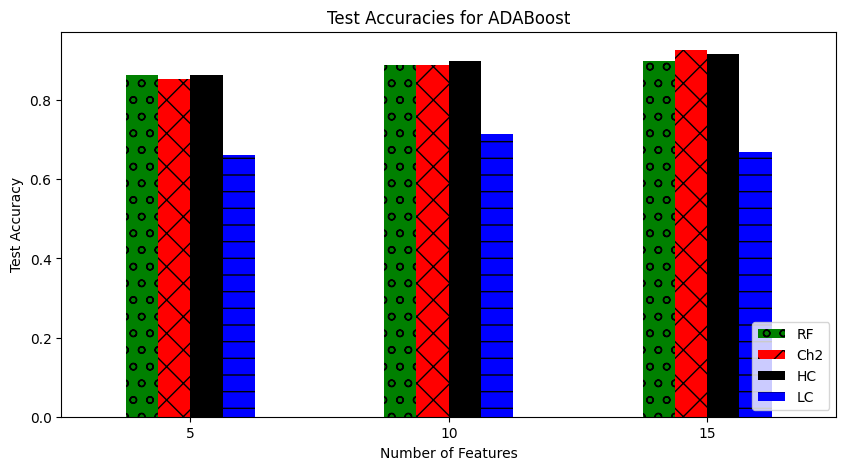}
  \label{fig54}}
  \hfil
  \subfloat[RF]{\includegraphics[width=0.45\columnwidth]{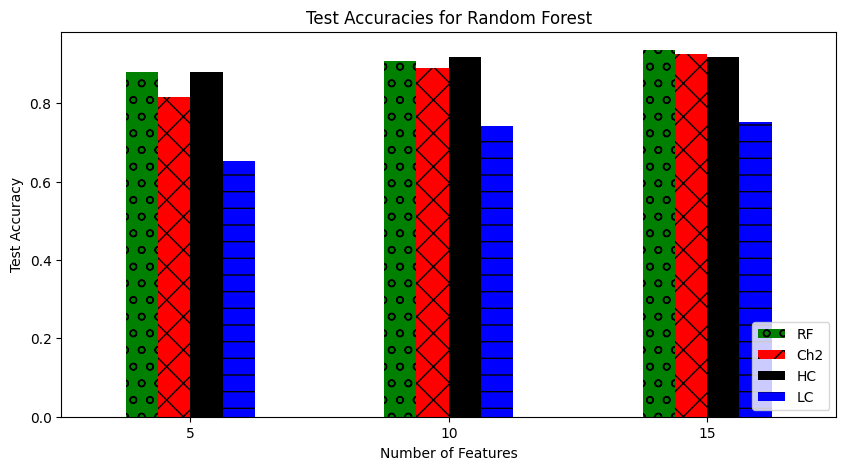}
  \label{fig55}}
  \caption{Test accuracies of oversampling for different ML algorithms.}
  \label{fig5}
\end{figure}
\subsection{Random Oversampling Minority Class}
In this study phase, we implemented the oversampling technique, a method that rebalances class distribution by replicating minority target instances randomly. The outcomes of this approach are presented in Table \ref{tab2}, which showcases the test accuracies obtained using CatBoost. The most remarkable achievement here is the highest test accuracy of 94.50\%, achieved using 15 features selected through HC. A comparative analysis of these outcomes can be visualized in Fig. \ref{fig51}. Equally, Table ~\ref{tab2} encapsulates the test accuracies for XGBoost, providing a comprehensive view of the performance, which can be compared visually using Fig. \ref{fig52}.

It is worth noting that the highest test accuracy consistently achieved was 91.74\%, obtained for both 10 and 15 features selected through RF feature importance and HC. This robust performance is evident across various models. In Table \ref{tab2}, the results for LGBM are delineated, showing that the highest test accuracy reached 93.58\% when 15 features were identified via Ch2. AdaBoost demonstrated exceptional performance with the highest test accuracy of 92.65\% when utilizing 15 features selected through Ch2, as observed in Table \ref{tab2} and Fig. \ref{fig54}. In conclusion, RF outperformed other algorithms, delivering a test accuracy of 93.58\% when 15 features were chosen based on RF feature importance, a result demonstrated in Table \ref{tab2} and Fig. \ref{fig55}. Notably, these algorithms exhibited remarkable performance, particularly when employing 15 features selected through RF feature importance.
\begin{table}[htbp]
\caption{Test accuracies of different algorithms with different features selection approaches for oversampling}
\begin{center}
\begin{tabular}{|c|c|c|c|c|}
\hline
\textbf{No. of Features} & \textbf{RF} & \textbf{Ch2} & \textbf{HC} & \textbf{LC} \\
\hline
\multicolumn{5}{|c|}{CatBoost}\\
\hline
5 & 87.16\% & 81.65\% & 87.16\% & 74.31\%\\
\hline
10 & 93.58\% & 92.66\% & 90.83\% & 77.98\%\\
\hline
15 & 91.74\% & 91.74\% & 94.50\% & 78.90\%\\
\hline
\multicolumn{5}{|c|}{XGBoost}\\
\hline
5 & 86.24\% & 82.57\% & 86.24\% & 63.30\%\\
\hline
10 & 91.74\% & 88.99\% & 89.91\% & 77.06\%\\
\hline
15 & 89.91\% & 90.83\% & 91.74\% & 77.98\%\\
\hline
\multicolumn{5}{|c|}{LGBM}\\
\hline
5 & 85.32\% & 82.57\% & 85.32\% & 64.42\%\\
\hline
10 & 92.66\% & 90.83\% & 89.91\% & 77.06\%\\
\hline
15 & 90.83\% & 93.58\% & 91.74\% & 79.82\%\\
\hline
\multicolumn{5}{|c|}{AdaBoost}\\
\hline
5 & 86.24\% & 85.32\% & 86.24\% & 66.06\%\\
\hline
10 & 88.99\% & 88.99\% & 89.91\% & 71.56\%\\
\hline
15 & 89.91\% & 92.65\% & 91.74\% & 66.97\%\\
\hline
\multicolumn{5}{|c|}{RF}\\
\hline
5 & 88.07\% & 81.65\% & 88.07\% & 65.14\%\\
\hline
10 & 90.83\% & 88.99\% & 91.74\% & 74.31\%\\
\hline
15 & 93.58\% & 92.66\% & 91.74\% & 75.23\%\\
\hline
\end{tabular}
\end{center}
\label{tab2}
\end{table}

\section{Conclusions}
This paper presents a data-driven approach to diagnosing PCOS in women, leveraging various techniques, including undersampling, oversampling, and powerful boosting algorithms for ML and feature selection. The dataset comprises 541 patients with 43 features and one target attribute, and it underwent rigorous feature selection to identify the 15 most crucial variables from the original 43. Notably, our results underscore the significance of feature selection, as our algorithms demonstrated enhanced performance when utilizing the top 10 features identified by RF. AdaBoost, in particular, achieved the highest test accuracy of 94.44\% when employing the 10 features selected by RF and HC. This highlights the pivotal role of feature choice and selection methodologies in shaping the performance of ML models. Future directions in this field involve exploring more interpretable models, collecting more extensive and diverse datasets, and integrating domain knowledge into the feature selection process further to enhance the accuracy and reliability of PCOS diagnosis. Additionally, addressing ethical considerations, such as algorithmic bias and transparency, should be a primary focus in developing medical AI systems.


\begin{thebibliography}{00}
\bibitem{b1} S. Ashraf, M. Nabi, S.u.A. Rasool, et al., "Hyperandrogenism in polycystic ovarian syndrome and role of CYP gene variants: a review," \textit{Egypt J Med Hum Genet}, vol. 20, p. 25, 2019.

\bibitem{b2} R. Pasquali, E. Stener-Victorin, B. O. Yildiz, et al., "Pcos forum: research in polycystic ovary syndrome today and tomorrow," \textit{Clinical endocrinology}, vol. 74, no. 4, pp. 424–433, 2011.

\bibitem{b3} A. S. Lagana, S. G. Vitale, M. Noventa, and A. Vitagliano, "Current management of polycystic ovary syndrome: from bench to bedside," \textit{International journal of endocrinology}, vol. 2018, 2018.

\bibitem{b4} S. A. Suha and M. N. Islam, "An extended machine learning technique for polycystic ovary syndrome detection using ovary ultrasound image," \textit{Scientific Reports}, vol. 12, no. 1, p. 17123, 2022.

\bibitem{b5} E. Setiawati, A. Tjokorda, W. Astuti, et al., "Particle swarm optimization on follicles segmentation to support PCOS detection," in \textit{2015 3rd international conference on information and communication technology (ICoICT)}. IEEE, 2015, pp. 369–374.

\bibitem{b6} B. Purnama, U. N. Wisesti, F. Nhita, A. Gayatri, T. Mutiah, et al., "A classification of polycystic ovary syndrome based on follicle detection of ultrasound images," in \textit{2015 3rd International Conference on Information and Communication Technology (ICoICT)}. IEEE, 2015, pp. 396–401.

\bibitem{b7} B. Purnama, A. Hasyim, M. Septiani, U. Wisesty, W. Astuti, et al., "Follicle detection on the USG images to support determination of polycystic ovary syndrome," in \textit{Journal of Physics: Conference Series}, vol. 622, no. 1. IOP Publishing, 2015, p. 012027.

\bibitem{b8} U. N. Wisesty, J. Nasri, et al., "Modified backpropagation algorithm for polycystic ovary syndrome detection based on ultrasound images," in \textit{International Conference on Soft Computing and Data Mining}. Springer, 2016, pp. 141–151.

\bibitem{b9} M. Maharani, B. Dewi, F. Yulianto, B. Purnama, et al., "Digital image compression using graph coloring quantization based on wavelet-svd," in \textit{Journal of Physics: Conference Series}, vol. 423, no. 1. IOP Publishing, 2013, p. 012019.

\bibitem{b10} E. Setiawati, T. A. Wirayuda, W. Astuti, et al., "A classification of polycystic ovary syndrome based on ultrasound images using supervised learning and particle swarm optimization," \textit{Advanced Science Letters}, vol. 22, no. 8, pp. 1997–2001, 2016.

\bibitem{b11} P. Mehrotra, C. Chakraborty, B. Ghoshdastidar, S. Ghoshdastidar, and K. Ghoshdastidar, "Automated ovarian follicle recognition for polycystic ovary syndrome," in \textit{2011 International Conference on Image Information Processing}. IEEE, 2011, pp. 1–4.

\bibitem{b12} S. Rihana, H. Moussallem, C. Skaf, and C. Yaacoub, "Automated algorithm for ovarian cysts detection in ultrasonogram," in \textit{2013 2nd International Conference on Advances in Biomedical Engineering}. IEEE, 2013, pp. 219–222.

\bibitem{b13} J. Canny, "A computation approach to edge detection," \textit{IEEE Trans. Pattern Anal. Mach. Intell.}, vol. 8, no. 6, pp. 670–700, 1986.

\bibitem{b14} “Pcos diagnosis — kaggle,” https://www.kaggle.com/code/karnikakapoor/ pcos-diagnosis/data, (Accessed on 12/05/2022).

\bibitem{b15} "How to fill nan values with mean in pandas? - geeksforgeeks," https://www.geeksforgeeks.org/how-to-fill-nan-values-with-mean-in-pandas/, (Accessed on 12/03/2022).

\bibitem{b16} “Normalization — codecademy,” https://www.codecademy.com/article/ normalization, (Accessed on 12/03/2022).

\bibitem{b17} J. Brownlee, "Random oversampling and undersampling for imbalanced classification," \textit{Machine Learning Mastery}, 2020.

\bibitem{b18} D. Elreedy and A. F. Atiya, "A comprehensive analysis of synthetic minority oversampling technique (SMOTE) for handling class imbalance," \textit{Information Sciences}, vol. 505, pp. 32-64, 2019.

\bibitem{b19} L. Prokhorenkova, G. Gusev, A. Vorobev, A. V. Dorogush, and A. Gulin, "CatBoost: unbiased boosting with categorical features," in \textit{Advances in neural information processing systems}, vol. 31, 2018.

\bibitem{b20} D. Zhang, L. Qian, B. Mao, C. Huang, B. Huang, and Y. Si, "A data-driven design for fault detection of wind turbines using random forests and xgboost," \textit{Ieee Access}, vol. 6, pp. 21 020–21 031, 2018.

\bibitem{b21} G. Ke, Q. Meng, T. Finley, T. Wang, W. Chen, W. Ma, Q. Ye, and T.Y. Liu, "Lightgbm: A highly efficient gradient boosting decision tree," in \textit{Advances in neural information processing systems}, vol. 30, 2017.

\bibitem{b22} X. Li, L. Wang, and E. Sung, "Adaboost with SVM-based component classifiers," \textit{Engineering Applications of Artificial Intelligence}, vol. 21, no. 5, pp. 785–795, 2008.

\bibitem{b23} M. Pal, "Random forest classifier for remote sensing classification," \textit{International journal of remote sensing}, vol. 26, no. 1, pp. 217–222, 2005.

\bibitem{b24} "How does feature selection benefit machine learning tasks?" https://h2o.ai/wiki/feature-selection/, (Accessed on 12/03/2022).

\bibitem{b25} A. Dubey, "Feature selection using random forest," \textit{Towards Data Science}, 2018.

\bibitem{b26} S. K. Gajawada, "Chi-square test for feature selection in machine learning," \textit{Towards Data Science}, 2019.

\bibitem{b27} R. Vishal, "Feature selection-correlation and p-value," \textit{Towards Data Science}, vol. 20, p. 20, 2018.

\end{thebibliography}
\end{document}